\documentclass[10pt,twocolumn,letterpaper]{article}

\usepackage[pagenumbers]{wacv} % To force page numbers, e.g. for an arXiv version

\usepackage[table]{xcolor}

\definecolor{wacvblue}{rgb}{0.21,0.49,0.74}
\usepackage[pagebackref,breaklinks,colorlinks,allcolors=wacvblue]{hyperref}
\usepackage{comment}

\def\wacvPaperID{*****} % *** Enter the WACV Paper ID here
\def\confName{WACV}
\def\confYear{2026}

\title{TrashDet: Iterative Neural Architecture Search for Efficient Waste Detection}

\author{Tony Tran\\
Department of Research Computing\\
University of Houston\\
Houston, TX 7704\\
{\tt\small thtran37@cougarnet.uh.edu}
\and
Bin Hu\\
Department of Engineering Technology\\
University of Houston\\
Houston, TX 77004\\
{\tt\small bhu11@central.uh.edu}
}

\begin{document}
\maketitle
\begin{abstract}
This paper addresses trash detection on the TACO dataset under strict TinyML constraints using an iterative hardware-aware neural architecture search framework targeting edge and IoT devices. The proposed method constructs a Once-for-All-style ResDets supernet and performs iterative evolutionary search that alternates between backbone and neck/head optimization, supported by a population passthrough mechanism and an accuracy predictor to reduce search cost and improve stability. This framework yields a family of deployment-ready detectors, termed TrashDets. On a five-class TACO subset (paper, plastic, bottle, can, cigarette), the strongest variant, TrashDet-l, achieves \textbf{19.5 mAP50} with \textbf{30.5M} parameters, improving accuracy by up to \textbf{3.6 mAP50} over prior detectors while using substantially fewer parameters. The TrashDet family spans \textbf{1.2M} to \textbf{30.5M} parameters with mAP50 values between \textbf{11.4} and \textbf{19.5}, providing scalable detector options for diverse TinyML deployment budgets on resource-constrained hardware. On the MAX78002 microcontroller with the TrashNet dataset, two specialized variants, \textbf{TrashDet–ResNet} and \textbf{TrashDet–MBNet}, jointly dominate the ai87-fpndetector baseline, with TrashDet–ResNet achieving \textbf{7{,}525~$\mu$J} energy per inference at \textbf{26.7~ms} latency and \textbf{37.45} FPS, and TrashDet–MBNet improving mAP50 by \textbf{10.2\%}; together they reduce energy consumption by up to \textbf{88\%}, latency by up to \textbf{78\%}, and average power by up to \textbf{53\%} compared to existing TinyML detectors.

\end{abstract}
    
\section{Introduction}
\label{sec:intro}

Uncontrolled waste disposal and littering contribute directly to environmental pollution and degradation of public spaces. Automatic image based waste detection and classification enables large scale monitoring in settings such as streets, parks, rivers, and coastlines~\cite{olawade2024smartwaste}. In practice, these systems must operate continuously on battery powered or solar powered nodes, where network connectivity is unreliable and maintenance is costly. This makes real time detection on devices with very limited memory, compute, and energy a central requirement rather than an implementation detail~\cite{mittal2024lightweight,surantha2025realtime}.

Datasets such as TACO (Trash Annotations in Context)~\cite{taco2020} provide realistic benchmarks for this problem, with cluttered backgrounds, small and deformable objects, and strong class imbalance across waste categories. Recent methods designed for solid waste and aerial litter, including SWDet and AltiDet~\cite{zhang2022swdet,LIEW2025110814}, improve detection quality on TACO and related datasets, but rely on backbones with tens of millions of parameters and GPU class hardware. Directly deploying such models on TinyML platforms such as microcontrollers is infeasible due to strict limits on convolutional operators, activation memory, and total layer count.

\begin{figure}[t]
  \centering
  \includegraphics[width=\linewidth]{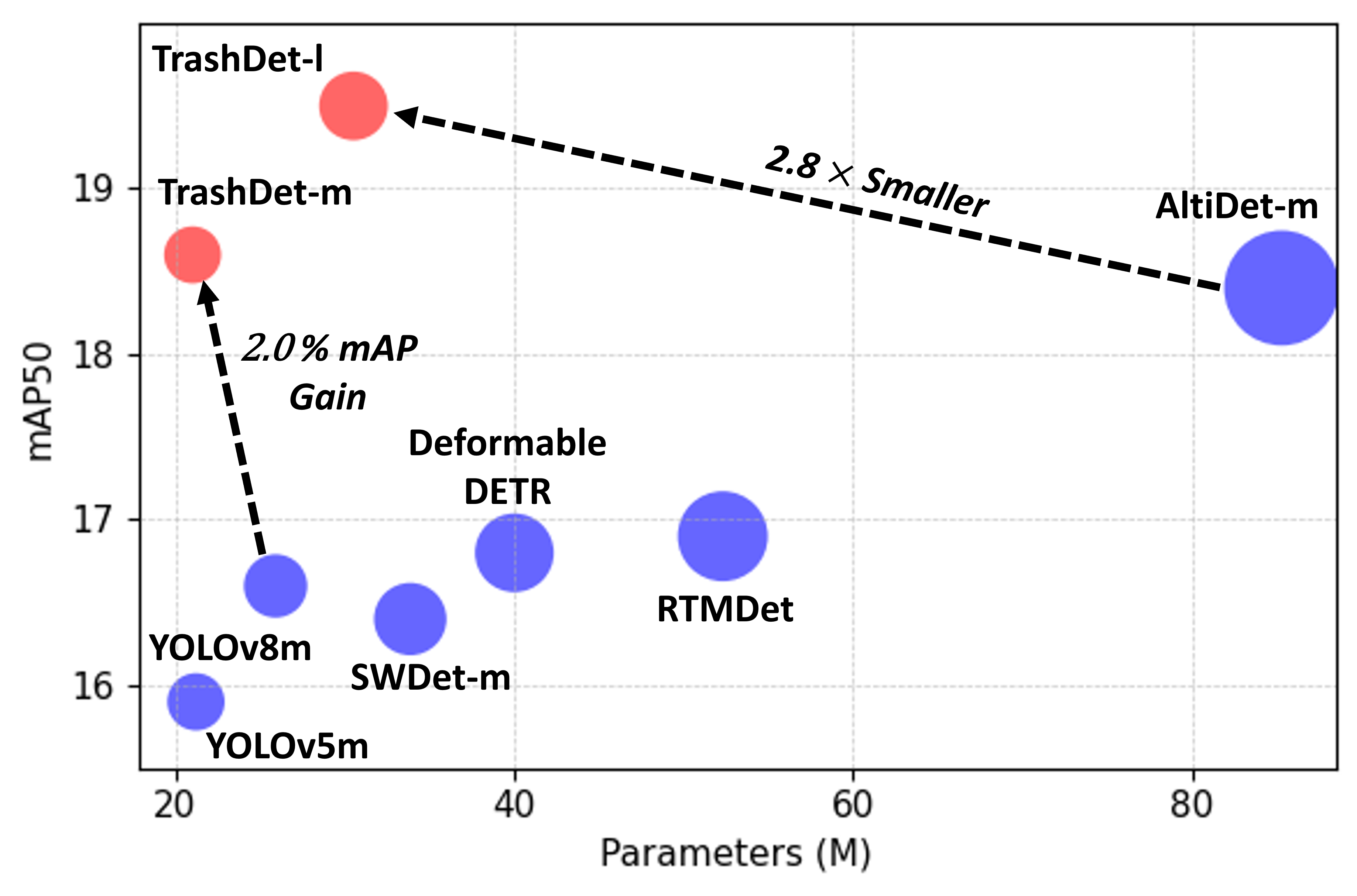}
  \caption{Comparisons with other state-of-the-art methods on the TACO~\cite{taco2020} dataset in terms of size-accuracy, reducing model size by 2.8x for similarly performing models and increasing accuracy by 2.0\% against similarly sized models.}
  \label{fig:topright}
\end{figure}

To address this gap, we employ an iterative hardware-aware NAS framework for real-time waste detection on constrained edge devices~\cite{tran2025}. It combines a Once-For-All (OFA) style weight-sharing detection supernet with an evolutionary search that alternates between backbone and head optimization under explicit resource limits. Population passthrough and a lightweight accuracy predictor maintain stability and efficiency in the non-convex combinatorial design space. This process yields a family of deployment-ready detectors, termed \textbf{TrashDet}, tailored to TACO and TinyML hardware budgets.

The methodology consists of the following components:
\begin{itemize}
    \item \textbf{Hardware-Constrained Supernet:} We utilize a hardware-constrained OFA detection supernet for trash recognition on the TACO dataset. The design supports flexible depth, width, and expansion-ratio configurations while strictly enforcing edge-device requirements, including supported operators, flash and activation-memory limits, and other deployment constraints.

    \item \textbf{Modular Search Strategy:} We adopt the iterative evolutionary NAS procedure~\cite{tran2025}, alternating backbone and head optimization under separate computational budgets. This modular search formulation reduces the complexity of the overall design space while still enabling coupled backbone--head architectures suitable for microcontroller deployment.

    \item \textbf{Search Stabilization:} To improve convergence and reduce search cost, we employ population passthrough and a learned accuracy predictor. These components stabilize the alternating search process and significantly reduce dependence on repeated full mAP50 evaluations.

    %\item \textbf{Evaluation and Deployment:} We evaluate the resulting TrashDet models on a five-class subset of TACO (paper, plastic, bottle, can, cigarette). The largest variant, TrashDet-l, achieves 19.5~mAP50 with 30.5M parameters, improving accuracy by up to 3.6~mAP50 over prior detectors while using substantially fewer parameters, as shown in Figure~\ref{fig:topright}. The full TrashDet family spans 1.2M to 30.5M parameters with mAP50 values between 11.4 and 19.5, providing scalable options for diverse TinyML deployment budgets. In comparison, a 1.08M-parameter TrashDet instance achieves 7{,}525~$\mu$J per inference at 26.7~ms latency and 37.45~FPS, reducing energy by up to 54{,}476~$\mu$J, latency by up to 95.9~ms and power by up to 235.3~mW compared to existing TinyML detector.
    \item \textbf{Evaluation and Deployment:} We evaluate the resulting TrashDet models on a five-class subset of TACO (paper, plastic, bottle, can, cigarette). The largest variant, TrashDet-l, achieves 19.5~mAP50 with 30.5M parameters, improving accuracy by up to 3.6~mAP50 over prior detectors while using substantially fewer parameters, as shown in Figure~\ref{fig:topright}. The full TrashDet family spans 1.2M to 30.5M parameters with mAP50 values between 11.4 and 19.5, providing scalable options for diverse TinyML deployment budgets. On the MAX78002 microcontroller with the TrashNet dataset, two specialized variants, \textbf{TrashDet–ResNet} and \textbf{TrashDet–MBNet}, jointly outperform the ai87-fpndetector baseline: TrashDet–ResNet achieves 7{,}525~$\mu$J per inference at 26.7~ms latency and 37.45~FPS with 1.08M parameters, while TrashDet–MBNet attains 93.3~mAP50 at 17{,}581~$\mu$J and 51.1~ms, yielding energy, latency, and power reductions of up to 88\%, 78\%, and 53\%, respectively, compared to the baseline.
\end{itemize}

\begin{figure*}[t]
  \centering
  \includegraphics[width=\linewidth]{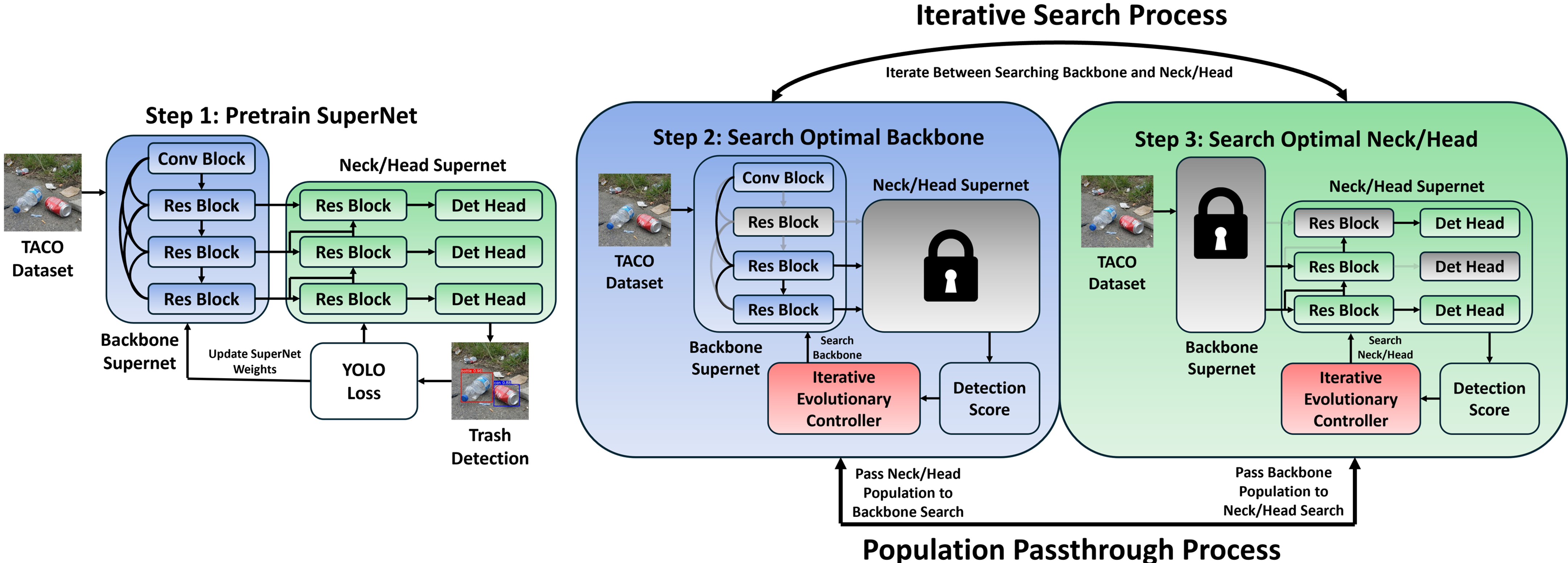}
  \caption{Overview of \textbf{TrashDet} framework. A unified OFA-style supernet is first constructed, comprising a ResNet-style backbone, neck, and YOLO-style detection head (left). Iterative evolutionary search is then performed in two coordinated stages: searching for the optimal backbone while keeping the neck/head fixed (Step~2), and searching for the optimal neck/head while fixing the discovered backbone (Step~3). The evolutionary controller evaluates candidate subnets using detection performance on TACO and alternates between these stages to obtain compact, deployment-ready TrashDet models.}
  \label{fig:methodw}
\end{figure*}

\section{Related Work}
\label{sec:related}

\subsection{Object Detection}

Modern object detection has progressed from two stage region based frameworks to one stage and transformer based detectors. Early methods such as R-CNN and Faster R-CNN~\cite{girshick2014rcnn,ren2015faster} used deep convolutional backbones with region proposals. Single stage detectors such as YOLO and SSD~\cite{redmon2016yolo,liu2016ssd} improved speed by predicting bounding boxes and class scores directly from dense feature maps. Later models including YOLOv3~\cite{redmon2018yolov3}, YOLOv5~\cite{jocher2020yolov5}, YOLOv8~\cite{jocher2023yolov8}, and RTMDet~\cite{lyu2022rtmdet} refine the architecture and training setup to improve the accuracy and speed trade off. Transformer based methods such as DETR and Deformable DETR~\cite{carion2020detr,zhu2021deformable} further simplify the detection pipeline by casting detection as a set prediction problem and using attention based feature aggregation.

\subsection{General Waste Detection}

Waste management, illegal dumping, and environmental pollution create a need for automatic systems that can detect and classify waste in real scenes. Recent work uses deep learning and computer vision to support waste sorting, litter detection, and contamination monitoring, but still reports challenges in data quality, domain shift, and deployment cost~\cite{olawade2024smartwaste}. Public datasets such as TACO (Trash Annotations in Context)~\cite{taco2020} offer in the wild images of litter across many outdoor settings and waste categories and are now standard benchmarks for trash detection and segmentation.

Several methods target waste detection in aerial and outdoor imagery. SWDet~\cite{zhang2022swdet} introduces an anchor based detector for solid waste in aerial images, with an Asymmetric Deep Aggregation backbone that improves multi scale feature representation for small and sparse trash instances. AltiDet~\cite{LIEW2025110814} adds altitude informed feature fusion and a high resolution feature extraction backbone to handle viewpoint and scale changes in UAV imagery. Many applications such as smart bins, UAV based monitoring, and low cost environmental sensors require real time inference on limited hardware. This combination of difficult visual conditions and strict resource limits motivates hardware aware methods for waste detection, including AltiDet~\cite{LIEW2025110814}, SWDet~\cite{zhang2022swdet}, and ELASTIC for microcontroller deployment~\cite{tran2025}.

\subsection{Neural Architecture Search}

Neural Architecture Search automates network design by searching over a predefined space of candidate architectures~\cite{elsken2019nas}. A NAS system defines a search space and a search strategy and optimizes a performance objective under given constraints. Early work focused on image classification, but later research extends NAS to detection and other dense prediction tasks and often adds hardware oriented objectives such as latency, floating point operations, or memory usage~\cite{liu2022cenas,gupta2024adasnas,mecharbat2023hytnas}.

For edge and TinyML settings, NAS can tailor architectures to specific devices and applications~\cite{mittal2024lightweight,liu2022cenas}. Hardware aware NAS for object detection in automotive and embedded scenarios~\cite{gupta2024adasnas} shows that jointly optimizing accuracy and resource use yields models that run efficiently under tight budgets. The search problem remains non convex and combinatorial, especially when including choices for backbone, neck, and head. This motivates search strategies that reduce search space size and cost while still respecting hardware constraints in realistic waste detection deployments.

\subsection{Once For All}

The OFA approach~\cite{cai2020once} reduces the cost of hardware aware NAS by decoupling supernet training from architecture selection. Instead of training separate networks for many candidate architectures, OFA trains one overparameterized supernetwork that supports a wide set of sub networks with different depths, widths, kernel sizes, and input resolutions. A progressive shrinking schedule trains larger and smaller sub networks together so that all share weights effectively~\cite{cai2020once}. This allows fast evaluation and deployment of many specialized models without retraining from scratch.

OFA style methods are useful in waste detection on constrained devices, where users may need a range of models from very small networks for microcontrollers to larger networks for more capable edge hardware~\cite{tran2025}. By building an OFA detection supernet with a ResNet backbone~\cite{he2016deep}, multi scale feature aggregation via FPN and PAN~\cite{lin2017feature,liu2018path}, and a YOLO style detection head~\cite{redmon2018yolov3}, one can obtain a family of architectures that match both the visual complexity of trash detection and the operator and memory limits of embedded CNN accelerators.

\section{Problem Formulation}
\label{formulation}

We aim to design an efficient object detection model for trash recognition that maximizes accuracy on the TACO dataset while satisfying strict hardware constraints required for edge deployment. Let a candidate detection network be denoted by 
$
\mathcal{N}(f, \mathbf{W})
$
where $f$ represents the architectural configuration sampled from a weight sharing supernet, and $\mathbf{W}$ denotes the shared network parameters. The search space $\mathcal{A}$ contains all valid architectures.

\paragraph{Supernet Training.}  
Following the standard OFA framework, we first train a weight sharing supernet that provides reliable performance estimates for any subnet sampled from $\mathcal{A}$. Instead of solving a bi-level optimization problem, the supernet parameters are optimized by minimizing the expected training loss over randomly sampled architectures:
\begin{equation}
\mathbf{W}^* = 
\arg\min_{\mathbf{W}} \ 
\mathbb{E}_{f \sim \mathcal{A}}
\left[
    \mathcal{L}_{\text{train}}(\mathcal{N}(f, \mathbf{W}))
\right]
\label{eq:supernet_training_corrected}
\end{equation}
This objective is highly non-convex due to weight sharing, stochastic sampling, and interference between subnets. As a result, the optimization process does not guarantee convergence to a global minimum and may exhibit slow or unstable training dynamics.

\paragraph{Architecture Search.}
Given the trained supernet parameters $\mathbf{W}^*$, the goal of NAS is to identify the optimal architecture
\begin{equation}
f^* = 
\arg\min_{f \in \mathcal{A}} 
    \mathcal{L}_{\text{val}}(\mathcal{N}(f, \mathbf{W}^*))
\label{eq:single_level_search}
\end{equation}
\begin{equation}
\text{s.t.} \quad \text{Cost}(f) \le \tau
\label{eq:single_level_constraint}
\end{equation}
The search problem is further complicated by the fact that NAS operates over a discrete and inherently combinatorial architectural space, where depth choices, width settings, kernel sizes, and expansion ratios form a large collection of integer-valued decisions. As a consequence, the search cannot rely on standard gradient-based optimization and typically requires heuristic strategies such as evolutionary algorithms, sampling-based search, or differentiable relaxations. These heuristics are sensitive to initialization, stochasticity, and sampling variance, and they offer no optimality guarantees.

\paragraph{Iterative Decomposition of the Search Problem.}
Because the search space grows combinatorially when jointly optimizing the backbone, neck, and head, we decompose the architecture into two components:
\[
f = (b, h)
\]
where $b$ denotes the backbone configuration and $h$ denotes the head configuration (including the neck).

At iteration $t$, we first optimize the backbone with the head fixed:
\begin{equation}
b^{(t+1)} =
\arg\min_{b}
    \mathcal{L}_{\text{val}}(\mathcal{N}((b, h^{(t)}), \mathbf{W}^*))
\label{eq:iter_backbone}
\end{equation}
\begin{equation}
\text{s.t.} \quad \text{Cost}(b) \le \tau_b
\label{eq:iter_backbone_constraint}
\end{equation}

Next, we fix the backbone and optimize the head:
\begin{equation}
h^{(t+1)} =
\arg\min_{h}
    \mathcal{L}_{\text{val}}(\mathcal{N}((b^{(t+1)}, h), \mathbf{W}^*))
\label{eq:iter_head}
\end{equation}
\begin{equation}
\text{s.t.} \quad \text{Cost}(h) \le \tau_h
\label{eq:iter_head_constraint}
\end{equation}

The budgets must satisfy the inequality
\[
\tau_b + \tau_h \le \tau
\]
ensuring compliance with the overall hardware constraint.

This iterative formulation reduces the effective dimensionality of the overall combinatorial search space, making heuristic optimization more tractable in practice. Although the optimization process remains non-convex and heuristic-driven, the reduced search space allows the algorithm to efficiently explore high-performing architectures while ensuring that the final model satisfies the strict resource and operator constraints required for real-time trash detection.

\section{Proposed Method}
\label{sec:method}

%Figure~\ref{fig:methodw} presents an overview of the \textbf{TrashDet} framework. On the left, a unified OFA-style supernet is constructed that jointly parameterizes a ResNet-style backbone, a lightweight neck, and a YOLO-style detection head, exposing architectural choices (e.g., depth, width, and kernel size) across all three components. The middle and right parts of the figure then depict the iterative evolutionary search procedure, which operates in two coordinated stages: in Step~2, the controller searches for the optimal backbone by sampling and evaluating candidate backbone subnets while keeping the neck and head fixed; in Step~3, the discovered backbone is fixed and the controller instead searches over neck/head configurations. At each stage, candidate subnets are evaluated on TACO under device-level resource constraints, and high-performing architectures are propagated across iterations. By alternating between these two stages, the framework progressively refines both backbone and neck/head, ultimately producing compact, deployment-ready \textbf{TrashDet} models that achieve strong detection performance on TACO within strict TinyML budgets.

Figure~\ref{fig:methodw} presents an overview of the proposed \textbf{TrashDet} framework to address efficient waste detection on constrained edge devices. We first construct a unified OFA-style detection supernet that jointly parameterizes a ResNet-style backbone, a lightweight neck, and a YOLO-style detection head. An iterative evolutionary search then alternates between optimizing the backbone and the neck/head under device-level resource constraints, ultimately yielding compact, deployment-ready \textbf{TrashDet} models for TACO within strict TinyML budgets.

\subsection{Iterative Evolutionary Architecture Search}
\label{iterativeea}

Formally, to solve the alternating minimization problem defined in Section~\ref{formulation}, we employ an \textit{Iterative Evolutionary Architecture Search}. Unlike conventional evolutionary NAS, which attempts to optimize the entire architecture vector in a single global pass, our framework adopts a cyclic coordinate–descent strategy that alternates the evolutionary search between the backbone $b$ and the head $h$ over successive iterations $t$. This modular decomposition reduces the effective search space dimensionality, enabling coordinated adaptation of architectural components under strict TinyML constraints.

For a given module (e.g., the backbone $b$) at iteration $t$, the evolutionary procedure proceeds as follows:

\begin{itemize}
    \item \textbf{Warm-start Initialization}:  
    Instead of sampling a population from scratch, the initial population $\mathcal{P}_0^{(t)}$ is seeded with the top-$k$ highest-performing architectures from iteration $t-1$. This warm-starting strategy preserves optimization progress, stabilizes convergence, and refines strong candidates across iterations.

    \item \textbf{Constraint-Aware Variation}:  
    Offspring are generated via mutation and crossover. Mutation perturbs architectural hyperparameters (e.g., depth, width, kernel size), while crossover exchanges structural components between parents. Before evaluation, each offspring undergoes strict hardware-feasibility checks; any candidate violating the module budget (e.g., $\text{Cost}(b) > \tau_b$) is discarded or re-sampled so no computation is wasted on non-deployable architectures.

    \item \textbf{Proxy Fitness Evaluation}:  
    Valid candidates are instantiated within the weight-sharing supernet $\mathcal{N}$ and evaluated on a TACO validation subset under device-level resource constraints. The fitness measures the contribution of the candidate module to the full detector, combined with the complementary fixed module. This proxy evaluation avoids retraining and provides fast, reliable performance estimates.

    \item \textbf{Tournament Selection}:  
    Tournament selection determines which candidates survive to the next generation, imposing selection pressure toward architectures that maximize detection accuracy while complying with the allocated resource budget.
\end{itemize}

Overall, the search follows the three-step loop illustrated in Figure~\ref{fig:methodw}. \textbf{Step~1} constructs the unified OFA-style supernet that jointly parameterizes the backbone, neck, and head. \textbf{Step~2} applies the evolutionary cycle above to the backbone, with the neck/head fixed, yielding an updated backbone $b^{(t)}$. \textbf{Step~3} then freezes $b^{(t)}$ and runs the same evolutionary procedure on the neck/head to obtain $h^{(t)}$. This three-step process is repeated for multiple global iterations, alternately refining $b$ and $h$ while respecting hardware budgets, and converging to compact, deployment-ready \textbf{TrashDet} architectures for TinyML devices.

\subsection{Population Passthrough}
\label{passthrough}

A challenge inherent to alternating modular search is the potential loss of useful search information when transitioning between modules. In standard evolutionary optimization, switching focus from one module (e.g., backbone $b$) to another (e.g., head $h$) often triggers a full population reinitialization, which disrupts the optimization trajectory and reintroduces redundant exploration. To improve search stability and preserve continuity across module alternations, we employ a \textit{Population Passthrough} mechanism that retains a controlled fraction of high-quality candidates for each module.

Each module $m \in \{b, h\}$ maintains a memory buffer $\mathcal{M}_m$ that stores its highest-performing architectures from previous optimization cycles. When module $m$ is revisited at iteration $t$, the initial population is constructed using a fixed passthrough ratio $\rho \in [0,1]$ that determines what fraction of the population is retained from memory, while the remainder is filled by sampling new architectures from the supernet:

\begin{equation}
    \mathcal{P}^{(t)} 
    = 
    \underbrace{\text{Top}_{\rho N}(\mathcal{M}_m)}_{\text{Elite Passthrough}}
    \;\cup\;
    \underbrace{\text{Sample}\big(\mathcal{A}_m,\, N - \rho N\big)}_{\text{Diversity Augmentation}}
\end{equation}
\begin{itemize}
    \item $N$ is the total population size,
    \item $\rho$ is the passthrough ratio,
    \item $\text{Top}_{\lfloor \rho N \rfloor}(\mathcal{M}_m)$ returns the best-performing architectures stored in the memory buffer,
    \item $\text{Sample}(\mathcal{A}_m,\, \cdot)$ generates new feasible architectures from the module search space $\mathcal{A}_m$ while enforcing the constraint $\text{Cost}(m) \le \tau_m$.
\end{itemize}

This construction yields two complementary components:

\textbf{Elite Passthrough (Exploitation)}:  
A fraction $\rho$ of the population is inherited directly from previously validated high-performance architectures, ensuring that search progress is not lost during module alternation.

\textbf{Diversity Augmentation (Exploration)}:  
The remaining fraction $(1-\rho)$ is composed of newly sampled candidates, providing sufficient variability to explore novel architectural configurations that may become favorable as the complementary module evolves.

Unlike conventional elitism or aging mechanisms—which operate within a single continuous population—Population Passthrough functions across alternating optimization cycles. This inter-module retention is particularly beneficial in modular NAS, where the effective search landscape of one module depends on the evolving structure of the other, enabling more stable and efficient convergence under resource-constrained TinyML settings.

\section{Experimental Results and Analysis}
\label{sec:experiment}

\begin{figure}[t]
  \centering
  \includegraphics[width=\linewidth]{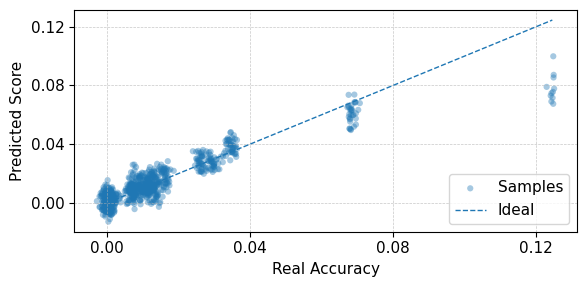}
  \caption{Accuracy predictor versus true mAP50 for candidate subnets. Points lying close to the dashed identity line indicate that the predictor is a reliable surrogate for detection performance during search.}
  \label{fig:predictor}
\end{figure}

In this section, we empirically evaluate our framework on the TACO dataset under both accuracy and hardware efficiency criteria. We first describe the experimental setup, including the dataset curation, evaluation metrics, and supernet design, as well as implementation details of the iterative evolutionary search and deployment constraints. We then compare our method against a range of state-of-the-art detectors, highlighting the accuracy–efficiency trade-offs achieved by our method. Finally, we analyze performance across different variants tailored to TinyML settings.

\subsection{Experimental Setup}

\subsubsection{Dataset}

For training, validation, and testing of our method, we adopt the Trash Annotations in Context (TACO) ~\cite{taco2020} dataset, a widely used benchmark for waste related computer vision research. TACO ~\cite{taco2020} comprises approximately 1,500 real world images collected in diverse outdoor environments and annotated across 60 waste categories. Owing to its realistic conditions, heterogeneous backgrounds, and mixture of small and deformable objects, the dataset has become a standard benchmark for litter detection, waste sorting automation, and environmental monitoring applications. These characteristics make TACO ~\cite{taco2020} well suited for assessing the robustness and generalization capability of detection architectures under practical deployment scenarios.

Although TACO ~\cite{taco2020} offers broad class diversity, it also exhibits a high degree of class imbalance. To ensure consistency and enable direct comparison with prior work, we curate a focused subset of TACO ~\cite{taco2020} containing only images corresponding to the paper, plastic, bottle, can, and cigarette classes. We therefore use this TACO ~\cite{taco2020} subset as our primary benchmark for evaluating waste-detection performance, analyzing the model performance and efficiency. This controlled yet realistic setting allows us to draw reliable conclusions about our method’s comparative effectiveness and its suitability for practical environmental-monitoring pipelines.

\subsubsection{Evaluation Metric}

Average Precision (AP) is the standard evaluation metric for object detection, computed as the area under the precision–recall curve and reflecting the trade-off between precision and recall across detection thresholds. In our experiments, we report mean Average Precision at an IoU threshold of 0.5 (mAP50) to assess the performance of each fine-tuned subnet. However, although mAP50 is widely used, it provides a coarse and computationally expensive feedback signal during neural architecture search. Evaluating mAP50 for every candidate architecture substantially increases search time and can impede efficient exploration of the search space.

To address this limitation, we incorporate an accuracy predictor that estimates the performance of a candidate subnet based on its expected mAP50 after the progressive shrinking algorithm ~\cite{cai2020once}. Rather than computing mAP50 directly at each search iteration, our predictor generates a lightweight performance estimate that guides the search dynamics. This strategy significantly reduces computational overhead while maintaining strong correlation with true downstream detection performance, enabling more efficient and scalable architecture optimization. Figure ~\ref{fig:predictor} illustrates the comparison between the actual mAP50 values and the outputs of our accuracy predictor, demonstrating the predictor’s ability to approximate true detection performance.

\begin{figure}[t]
  \centering
  \includegraphics[width=\linewidth]{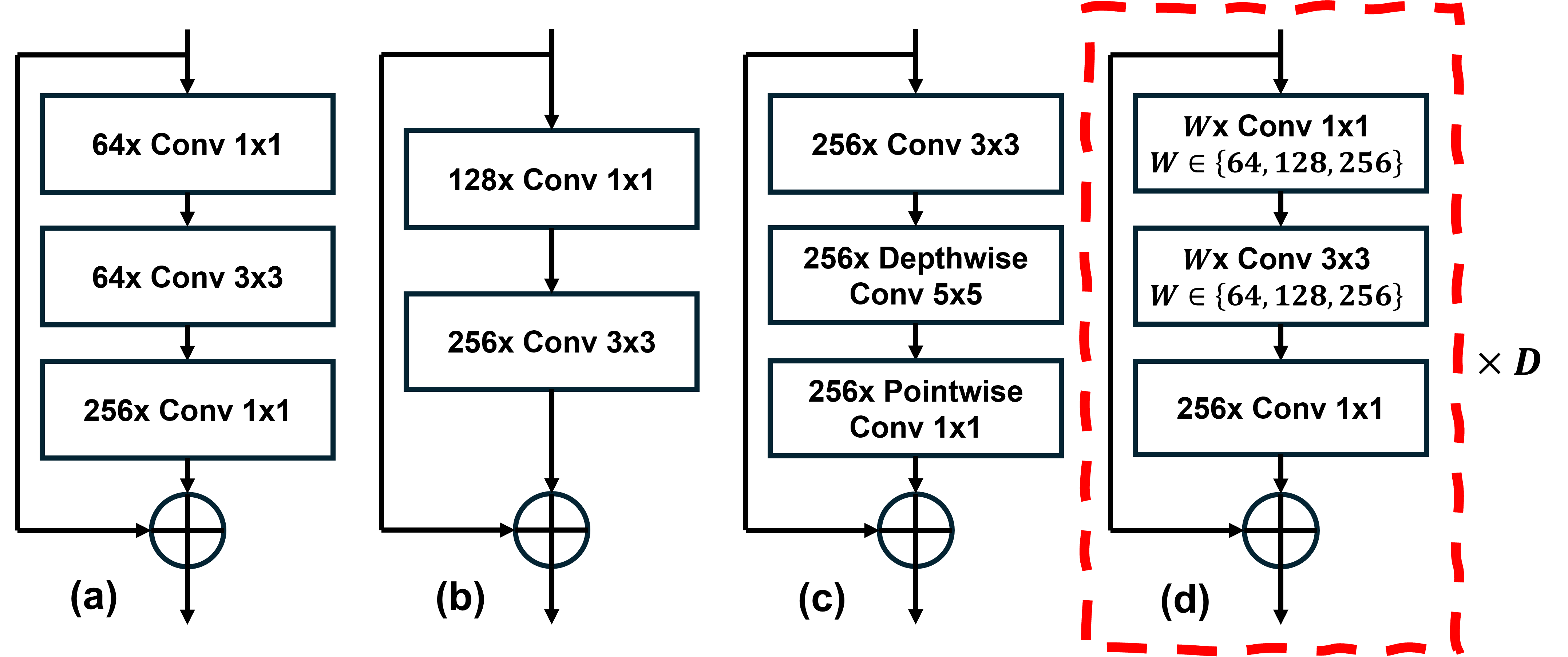}
  \caption{Basic Building Blocks. (a) The original bottleneck block used in ResNets~\cite{he2016deep}. (b) The DarkNet block used in CSPDarknets~\cite{wang2020cspnet}. (c) RepVGGBlock block used in RTMDet~\cite{lyu2022rtmdet}. (d) The OFA-Res Block Used in OFA ResNet~\cite{cai2020once}.}
  \label{fig:blocks}
\end{figure}

\subsubsection{Supernet Design}

We design an OFA ~\cite{cai2020once} ResDets ~\cite{he2016deep} supernetwork that unifies the backbone, neck, and detection head within a single search space. This search space supports dynamic specialization along three architectural dimensions: (1) depth (i.e., the number of active blocks per stage), (2) width (channel scaling), and (3) expansion ratio for each residual bottleneck block. The supernetwork consists of a dynamic OFA ResNet backbone, a dynamic FPN ~\cite{lin2017feature}, a dynamic PAN ~\cite{liu2018path}, and a dynamic YOLO detection head ~\cite{redmon2018yolov3}, all of which are parameterized to support multiple architectural configurations within a single over-parameterized model.

For each stage of the backbone, we vary the \textit{depth} by selecting between 2 and 8 residual blocks, allowing the network to adjust its representational capacity at different semantic levels. Each residual block supports multiple \textit{channel widths}, parameterized by a width-multiplier set
\[
\mathcal{W} = \{0.8,\ 1.0,\ 1.25,\ 1.5\}.
\]
All layers within the same stage share a common base width prior to scaling, ensuring consistent feature dimensionality across that stage.

In addition, each residual block is assigned an independent \textit{expansion ratio}, which determines the number of intermediate channels. We draw expansion ratios from
\[
\mathcal{E} = \{0.20,\ 0.25,\ 0.35,\ 0.45,\ 0.55\}.
\]

Beyond the backbone, we include fixed residual blocks whicih don't vary in depth within the FPN ~\cite{lin2017feature}, PAN~\cite{liu2018path}, and YOLO~\cite{redmon2018yolov3} detection head. These blocks participate in the search space through their width and expansion ratio choices. Figure~\ref{fig:blocks} demonstrates the differences between basic building blocks in previous methods against the one used in our experiments.

Because the backbone, neck, and head are jointly parameterized, the resulting search space is tightly coupled across all stages of the model. While this coupling significantly enlarges the total number of possible architectures, it also enables the search to discover globally optimized designs rather than isolated per-component improvements. To manage this combinatorial complexity, we employ an \textit{iterative search strategy} that efficiently explores high-performing subnets while avoiding the prohibitive cost of exhaustive evaluation.

\subsubsection{Implementation Details}

We perform iterative evolutionary search using standard evolutionary settings: an architectural mutation probability of 0.1, a population size of 100, a parent selection ratio of 0.25, and a mutation ratio of 0.5. For the iterative-search--specific configuration, we allocate a maximum budget of 10 iterations per module swap and allow up to 50 module swaps in total, with a population passthrough ratio of 0.5 to preserve high-performing candidates across iterations.

\begin{table*}[t]
  \centering
  \caption{Comparison of detectors on the TACO dataset. TrashDet-l achieves the highest mAP50 of 19.5 with 30.5M parameters, outperforming the strongest baseline AltiDet-m while using roughly one third of the parameters.}
  \label{tab:taco_results}
  \setlength{\tabcolsep}{6pt}
  \renewcommand{\arraystretch}{1.1}
  \begin{tabular}{lcccccc}
    \toprule
    Method & Backbone & Neck & Head & Params & AR & mAP50 \\
    \midrule
    YOLOv5m ~\cite{jocher2020yolov5}            & CSPDarknet~\cite{wang2020cspnet}   & SPPF~\cite{he2015spatial} + PANet~\cite{liu2018path}    & Yolov3~\cite{redmon2018yolov3}        & 21.2M     & \textbf{22.3} & 15.9 \\
    YOLOv8m ~\cite{jocher2023yolov8}            & CSPDarknet~\cite{wang2020cspnet}   & SPPF~\cite{he2015spatial} + PANet~\cite{liu2018path}    & Yolov8~\cite{jocher2023yolov8}        & 25.9M     & 16.6 & 16.6 \\
    \rowcolor{gray!15}
    \textbf{TrashDet-m (Ours)}                   & \textbf{OFA ResNet} & \textbf{OFA PANet} & \textbf{OFA Yolov3} & \textbf{21.0M}    & 19.1 & \textbf{18.6} \\
    \midrule
    SWDet-m ~\cite{zhang2022swdet}              & ADA~\cite{yu2018deep}           & EAFPN          & Yolov3~\cite{redmon2018yolov3}        & 33.85M    & 21.0 & 16.4 \\
    Deformable DETR ~\cite{zhu2021deformable}   & ResNet-101~\cite{he2016deep}    & DETR Encoder    & DETR Decoder~\cite{carion2020detr} & 40M        & \textbf{30.3} & 16.8 \\
    RTMDet ~\cite{lyu2022rtmdet}                & RTMDet-l       & PANet~\cite{liu2018path}           & RTMDet        & 52.3M     & 19.4 & 16.9 \\
    AltiDet-m ~\cite{LIEW2025110814}            & ADA + HRFE~\cite{wang2020deep}    & A-IFPN          & Yolov3~\cite{redmon2018yolov3}        & 85.3M     & 22.4 & 18.4 \\
    \rowcolor{gray!15}
    \textbf{TrashDet-l (Ours)}                   & \textbf{OFA ResNet} & \textbf{OFA PANet} & \textbf{OFA Yolov3} & \textbf{30.5M}   & 18.6 & \textbf{19.5} \\
    \bottomrule
  \end{tabular}
\end{table*}

\begin{table*}[t]
  \centering
  \caption{Performance of TrashDet variants on TACO. Model capacity ranges from TrashDet-n with 1.2M parameters to TrashDet-l with 30.5M parameters. Latency measurements were performed on an Ubuntu 22.04.4 system on an Intel Core i9-13900KF CPU and an NVIDIA GeForce RTX 4090 GPU.}
  \label{tab:variants}
  \setlength{\tabcolsep}{6pt}
  \renewcommand{\arraystretch}{1.1}
  \begin{tabular}{lcccccc}
    \toprule
    Method & Resolution & Params & AR & mAP50 & Latency (ms) & FPS \\
    \midrule
    TrashDet-n & 640 & 1.2M  & 21.2 & 11.4 & 2.21 & 452.79 \\
    TrashDet-s & 640 & 7.9M  & 16.9 & 15.8 & 3.83 & 261.06 \\
    TrashDet-m & 640 & 21.0M & 19.1 & 18.6 & 4.39 & 227.70 \\
    TrashDet-l & 640 & 30.5M & 18.6 & 19.5 & 5.07 & 197.08 \\
    \bottomrule
  \end{tabular}
\end{table*}

\subsubsection{Model Deployment}

We select one of our target deployment platform to be the \textbf{MAX78002}, a low-power microcontroller designed for energy constrained and low latency applications. The device integrates a hardware accelerated CNN accelerator with limited on-chip memory and tightly bounded compute resources, making it representative of real-time embedded inference scenarios. These characteristics align closely with our application requirements, where waste detection and classification must operate autonomously on battery powered or solar powered edge systems without access to high-performance compute resources. Consequently, our NAS framework must produce architectures that are not only accurate, but also highly compact and computationally efficient for practical deployment in environmental monitoring.

To properly integrate our neural network into the MAX78002 CNN accelerator, we must tightly adhere to the device's architectural constraints, as it does not support arbitrary network configurations in the same way that modern GPUs or general purpose microcontrollers do. For example, the accelerator imposes strict limitations on convolutional parameters such as kernel size, padding, and stride; only specific pooling and activation functions are supported; and both the input and output channel dimensions of any layer must not exceed 2048. The total number of layers that can be mapped to the accelerator is also capped, up to 128 CNN ``primal'' layers, and activation memory is restricted to approximately 80\,KiB.

In addition, we must account for the accelerator's \textit{streaming mode}, which allows intermediate activations to be reused across layers. Proper use of streaming mode enables more efficient memory utilization and permits higher input resolutions in the early stages of the network. Incorporating these hardware-aware considerations into the search process is essential for ensuring that the resulting architectures are feasible for deployment on the MAX78002 while still delivering strong real-time detection performance.

% \begin{table*}[t]
%     \centering
%     \small
%     \setlength{\tabcolsep}{6pt}
%     \renewcommand{\arraystretch}{1.1}
%     \caption{Energy, latency, and power comparison on the MAX78002. TrashDet reduces energy by up to 54{,}476~$\mu$J, latency by up to 95.9~ms, and power by up to 235.3~mW relative to ai87-fpndetector~\cite{ai85tinierssd}, and further lowers energy by 10{,}056~$\mu$J and latency by 24.4~ms compared to ELASTIC~\cite{tran2025}.}
%     \begin{tabular}{lccccccc}
%         \toprule
%         Model & Resolution & Device & Params & Energy ($\mu$J) & Latency (ms) & Power (mW) & FPS \\
%         \midrule
%         ai87-fpndetector~\cite{ai85tinierssd}
%             & 256$\times$320 & MAX78002 & 2.18M & 62001 & 122.6 & 445.76 & 8.16 \\
%         ELASTIC~\cite{tran2025} 
%             & 224$\times$224 & MAX78002 & 1.32M & 17581 & 51.1 & 285.02 & 19.57 \\
%         \rowcolor{gray!15}
%         \textbf{TrashDet (OURS)}
%             & 224$\times$224 & MAX78002 & \textbf{1.08M} & \textbf{7525} & \textbf{26.7} & \textbf{210.5} & \textbf{37.45} \\
%         \bottomrule
%     \end{tabular}
%     \label{tab:max78000_results}
% \end{table*}

\begin{table*}[t]
    \centering
    \small
    \setlength{\tabcolsep}{6pt}
    \renewcommand{\arraystretch}{1.1}
    \caption{Energy, latency, and power comparison on the MAX78002 for detectors discovered within the \textbf{TrashDet} search space. We explore ResNet- and MobileNet-style (MBNet) backbones and obtain two deployment-ready models, \textbf{TrashDet–MBNet} and \textbf{TrashDet–ResNet}, evaluated on the TrashNet dataset~\cite{aral2018classification} for detection. Compared to ai87-fpndetector~\cite{ai85tinierssd}, TrashDet reduces energy by up to 54{,}476~$\mu$J, latency by up to 95.9~ms, and power by up to 235.3~mW while maintaining superior accuracy.}
    \begin{tabular}{lcccccccc}
        \toprule
        Model & Resolution & Dataset & Params & Energy ($\mu$J) & Latency (ms) & Power (mW) & FPS & mAP50 \\
        \midrule
        ai87-fpndetector~\cite{ai85tinierssd}
            & 256$\times$320 & TrashNet & 2.18M & 62001 & 122.6 & 445.76 & 8.16 & 83.1 \\
        \textbf{TrashDet - MBNet}
            & 224$\times$224 & TrashNet & 1.32M & 17581 & 51.1 & 285.02 & 19.57 & \textbf{93.3} \\
        \textbf{TrashDet - ResNet}
            & 224$\times$224 & TrashNet & \textbf{1.08M} & \textbf{7525} & \textbf{26.7} & \textbf{210.5} & \textbf{37.45} & 84.6 \\
        \bottomrule
    \end{tabular}
    \label{tab:max78000_results}
\end{table*}

\subsection{Comparison Against State-of-the-Art Detectors}

Across the baselines, we observe a clear trade-off between model size, average recall (AR), and mAP50 on TACO in Table~\ref{tab:taco_results}. Deformable DETR~\cite{zhu2021deformable}, RTMDet-l~\cite{lyu2022rtmdet}, and AltiDet-m~\cite{LIEW2025110814} operate in a large-model regime with tens of millions of parameters and generally achieve higher AR, with Deformable DETR attaining the best recall with AR of 30.3 but only moderate mAP50 of 16.8. The YOLO-based detectors YOLOv5m~\cite{jocher2020yolov5} and YOLOv8m~\cite{jocher2023yolov8}, together with RTMDet-l~\cite{lyu2022rtmdet}, fall into a mid-range parameter budget from roughly twenty to fifty million parameters and achieve competitive but not leading performance, with mAP50 values clustered between 15.9 and 16.9. Among prior work, AltiDet-m~\cite{LIEW2025110814} stands out as the strongest baseline in terms of mAP50 with 18.4, but this improvement is accompanied by a substantially larger model with 85.3M parameters, which is undesirable for TinyML or edge deployment.

In contrast, TrashDet-l achieves the highest mAP50 on TACO with 19.5 while using only 30.46M parameters, outperforming all baselines in accuracy despite a substantially smaller footprint than AltiDet-m~\cite{LIEW2025110814} and even Deformable DETR~\cite{zhu2021deformable} and RTMDet-l~\cite{lyu2022rtmdet}. Although its AR of 18.6 is lower than the AR reported by Deformable DETR~\cite{zhu2021deformable} and AltiDet-m~\cite{LIEW2025110814}, the gain in mAP50 indicates that TrashDet-l produces more precise detections, emphasizing high-quality predictions over raw recall. This suggests that the architectures discovered by our NAS framework are better aligned with the long-tailed, cluttered characteristics of TACO, yielding a more favorable accuracy–efficiency trade-off that is particularly well suited to resource-constrained deployment scenarios.

\subsection{General Waste Detection on Tiny Devices}

To better understand the trade-off between accuracy and model capacity under TinyML constraints, we scale TrashDet into a family of architectures with varying parameter budgets, ranging from the ultra-compact TrashDet-n with 1.2M parameters to the larger TrashDet-l with 30.5M parameters, as summarized in Table~\ref{tab:variants}. All TACO variants share the same OFA Res backbone and input resolution, isolating the effect of capacity on detection performance and enabling practitioners to select an appropriate operating point for their target device and application constraints.

Beyond GPU evaluation, we adapt TrashDet for deployment on the MAX78002 microcontroller using the TrashNet dataset~\cite{aral2018classification}. Guided by the same hardware-aware search framework, we specialize the search space to explore both ResNet- and MobileNet-style (MBNet) backbones under the MAX78002 operator, activation memory, and layer-count budgets, yielding two deployment-ready subnets: \textbf{TrashDet–MBNet} and \textbf{TrashDet–ResNet}. Table~\ref{tab:max78000_results} highlightws the impact of these design choices, reporting energy, latency, power, FPS, and mAP50 alongside the existing ai87-fpndetector~\cite{ai85tinierssd} baseline.

Compared to ai87-fpndetector, TrashDet–ResNet reduces the energy per inference from 62{,}001~$\mu$J to 7{,}525~$\mu$J (\textbf{87.9\%} decrease) and lowers latency from 122.6~ms to 26.7~ms (\textbf{78.2\%} reduction). Average power drops from 445.76~mW to 210.5~mW, corresponding to a \textbf{52.8\%} decrease, while the parameter count is reduced from 2.18M to 1.08M (\textbf{50.5\%} fewer parameters). TrashDet–MBNet offers a complementary accuracy-focused design: it cuts energy from 62{,}001~$\mu$J to 17{,}581~$\mu$J (\textbf{71.6\%} reduction) and latency from 122.6~ms to 51.1~ms (a \textbf{58.3\%} reduction), while improving mAP50 from 83.1 to \textbf{93.3}, a gain of \textbf{10.2} points, at a substantially smaller model size (1.32M vs. 2.18M parameters).

Taken together, Table~\ref{tab:max78000_results} shows that, once the search space and hardware constraints are aligned with the MAX78002 accelerator, our framework can produce multiple Pareto-efficient detectors that dominate the hand-designed ai87-fpndetector across accuracy, energy, latency, and power. TrashDet–ResNet offers the most aggressive efficiency setting (37.45~FPS with 7{,}525~$\mu$J per inference), while TrashDet–MBNet delivers significantly higher accuracy at still dramatically lower cost than the baseline, making both models practical options for long-term battery-powered or energy-harvesting environmental monitoring at the edge.

% \begin{figure}[t]
%   \centering
%   \includegraphics[width=\linewidth]{imgs/max.jpg}
%   \caption{Review}
% \end{figure}
\section{Conclusion}
\label{sec:conclusion}

In this work, we utilize a hardware-aware neural architecture search framework to develop efficient and accurate object detectors for waste recognition under strict TinyML constraints. We combine an OFA-style weight-sharing supernet with an iterative evolutionary search that alternates between backbone and head optimization, using constraint-aware mutation, population passthrough, and a lightweight accuracy predictor. This enables exploration of a large architectural space while ensuring all discovered models remain deployable.

Our contributions are threefold. First, we construct a unified OFA ResDets supernetwork spanning depth, width, and expansion-ratio choices across the backbone, neck, and head. Second, we develop an iterative evolutionary search with population passthrough that stabilizes modular optimization under device constraints. Third, we demonstrate practical impact by discovering models from 1.2M to 30.5M parameters that outperform existing baselines on TACO, and by deploying the ultra-compact \textbf{TrashDet} model with significant reductions in energy, latency, and power.

By enabling real-time, low-power trash detection on microcontrollers, TrashDets supports scalable and long-term waste monitoring in outdoor and resource-limited settings. This contributes to cleaner public spaces, improved environmental stewardship, and broader societal impact through sustainable TinyML-enabled sensing.

{
    \small
    \bibliographystyle{ieeenat_fullname}
    \bibliography{main}
}

\end{document}